\documentclass{article} 
\usepackage{iclr2027_conference,times}

\usepackage{amsmath,amsfonts,bm}

\def\eqref#1{equation~\ref{#1}}

\def\1{\bm{1}}

\DeclareMathAlphabet{\mathsfit}{\encodingdefault}{\sfdefault}{m}{sl}
\SetMathAlphabet{\mathsfit}{bold}{\encodingdefault}{\sfdefault}{bx}{n}

\usepackage{hyperref}
\usepackage{url}

\usepackage{graphicx}
\usepackage{booktabs}
\usepackage{amsmath}
\usepackage{multirow}
\usepackage{threeparttable}
\usepackage{colortbl}
\usepackage{xcolor}

\usepackage{caption}
\definecolor{cellgreen}{HTML}{C5EFCE}
\definecolor{fontgreen}{HTML}{026102}

\title{How Latent Transition Geometry Matters \\for Planning in World Models}

\author{Xuanya Zhu$^{1}$, Da Li$^{2}$, Xiatian Zhu$^{1}$ \\
\\
$^{1}$ University of Surrey\\
$^{2}$ University of York\\
}

\iclrfinalcopy 
\begin{document}

\maketitle

\begin{abstract}
World models support planning by predicting how the environment evolves under candidate actions. 
However, one-step predictive accuracy alone does not characterize multi-step planning utility, which depends on distinguishing candidate futures.
This raises a natural question: \emph{How does latent transition geometry shape planning?} 
We investigate this through an action-conditioned transition-field view, using curvature regularization as a controlled intervention.
Across four continuous-control tasks, appropriately tuned regularization improves planning, while excessive regularization can be detrimental.
By decomposing the transition field into affine and non-affine components, we find that regularization reduces action sensitivity mainly through the non-affine component.
To assess its role in planning, we directly manipulate the non-affine action response in frozen models and show that it affects candidate selection and closed-loop success. 
Further analysis shows that curvature generally brings predicted futures closer together, while preserving more of the differences along the goal direction. 
This helps explain why the planning effects vary across tasks.
We observe broadly consistent planning trends across five sampling- and gradient-based solvers on LeWM and similar rollout-geometry trends on DINO-WM. 
Qualitative attention visualizations reveal accompanying shifts toward task-relevant visual structure. 
Together, these findings clarify how latent transition geometry influences multi-step planning.
\end{abstract}

\section{Introduction}
World models aim to learn predictive representations of environment dynamics for planning and control~\citep{ha2018recurrent,shen2026world}. Joint-embedding predictive architectures (JEPAs) provide a direct approach by predicting future states in latent space without reconstructing observations~\citep{lecun2022path}. In a latent world model, an encoder maps observations to latent states, and a predictor is trained to predict the next latent state from the current state and action. At deployment, a planner evaluates candidate action sequences by rolling this predictor forward and comparing their predicted futures.
This creates a gap between how the model is trained and how it is used for planning~\citep{jena2026world}. 
One-step predictive accuracy alone does not characterize planning utility~\citep{lambert2020objective,nair2020goal}, since planning depends on whether the distinctions among candidate futures remain informative after multi-step propagation. We illustrate this mismatch in Figure~\ref{fig:prediction_planning_gap} in the Appendix~\ref{app:pred_plan_gap}.

Recent work has addressed several aspects of this problem. Theoretical analyses characterize when JEPA representations recover latent world variables~\citep{klindt2026does} and when controlled dynamics are identifiable~\citep{zhang2026identifiability}. Other work connects pretraining error to downstream planning regret~\citep{cui2026generalization} and empirically investigates how model and planner choices affect physical planning performance~\citep{terver2025drives}. More recently, latent geometry has emerged as another perspective on planning, with curvature regularization showing promising improvements in robotic control~\citep{wang2026temporal}. Building on this direction, we ask a more general question: \emph{How does latent transition geometry shape planning?}

Specifically, we study this question in LeWorldModel (LeWM)~\citep{maes2026leworldmodel}, which jointly learns a latent representation and an action-conditioned predictor. We view predicted latent displacements as a \emph{transition field}, tracing how action-induced one-step differences evolve through rollouts into the candidate futures compared by the planner. 
We use the aforementioned curvature regularization as a controlled intervention on this geometry. Across TwoRoom~\citep{sobal2026learning}, PushT~\citep{chi2025diffusion}, Reacher~\citep{tassa2018deepmind}, and Cube~\citep{park2025ogbench}, appropriately tuned regularization improves planning, while excessive regularization can become detrimental.

\begin{figure}[t]
\centering
\includegraphics[width=\linewidth]{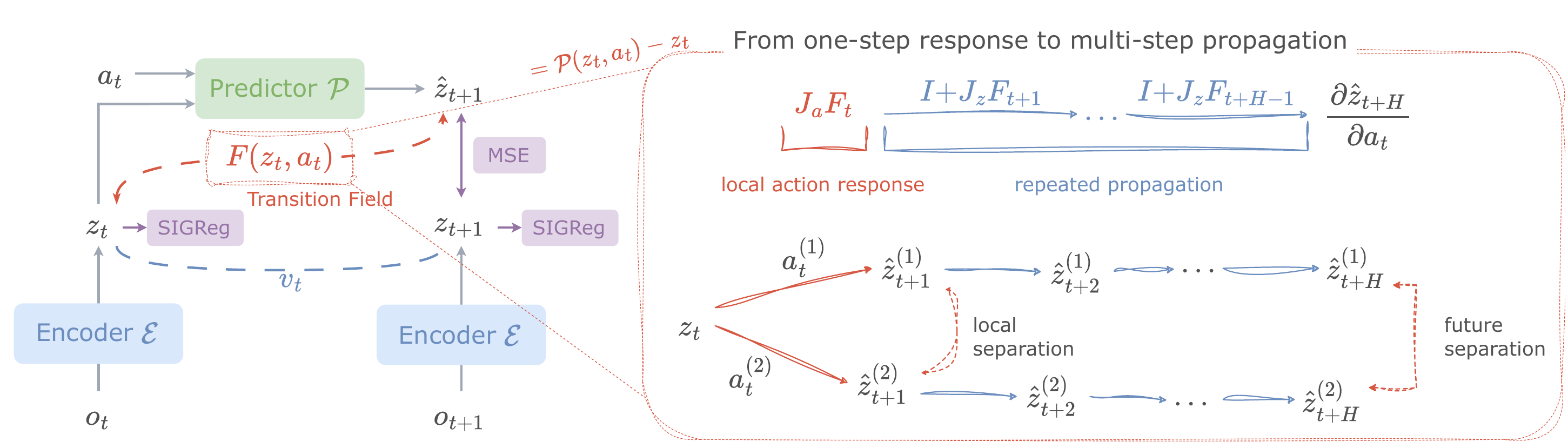}
\caption{Overview of the transition field. LeWM encodes observations into latent states and predicts the latent change produced by an action, which we denote by the transition field \(F(z_t,a_t)\). During planning, we repeatedly compose these predicted changes along candidate action sequences. Differences introduced by actions at one step are therefore transformed over later rollout steps, determining how separated the candidate futures remain at the planning horizon.}
\label{fig:main}
\end{figure}

To understand these outcomes, we examine how curvature reshapes the action-conditioned transition field. A concurrent work characterises controlled world models theoretically under linear-Gaussian latent dynamics, where the identified transition takes an affine state-action form~\citep{zhang2026identifiability}. 
Using this affine form as a reference, we decompose the field into an affine component and a non-affine residual to assess how much of its action response the reference explains and whether it captures curvature-induced changes.
We find that the non-affine component accounts for most of the reduction in one-step action-response energy. To test its role in planning, we rescale its action-induced differences while keeping the model weights frozen. This changes the quality of the action sequences selected by CEM~\citep{deboer2005tutorial} and affects closed-loop success, with effects that vary across tasks and checkpoints.

To understand what makes these differences useful for planning, we examine how they propagate to the planning horizon. As regularization strengthens, the overall rollout response to small action changes generally contracts, while a larger share of the remaining response lies along the goal direction. Even as predicted futures become more compact, nearby action sequences can still yield distinct predicted distances to the goal, helping the planner compare them. Excessive contraction, however, may weaken these distinctions and reduce planning performance. At the representation level, encoder attention maps reveal accompanying shifts in visual emphasis across these regimes.

Our contributions are threefold.
\emph{First}, we formulate latent prediction as an action-conditioned transition field and use it to trace how action-induced one-step differences propagate through multi-step rollouts.
\emph{Second}, we show that curvature reduces one-step action sensitivity mainly through the non-affine component, and that directly scaling its action-induced differences in frozen models affects CEM candidate selection and closed-loop success.
\emph{Third}, we characterize overall rollout contraction and relative goal-direction retention, relating these changes to task-dependent planning outcomes and complementary encoder-attention observations. 
We extend the evaluation to five planning solvers on LeWM and reproduce the rollout-geometry trends with DINO-WM.

\section{Related Work}
\noindent\textbf{World models} 
learn environment dynamics for planning, control, and simulation~\citep{ha2018recurrent}. Latent approaches such as PlaNet~\citep{hafner2019learning} and the Dreamer family~\citep{hafner2019dream} support planning or policy learning through imagined rollouts, while subsequent work explores discrete autoregressive dynamics with IRIS~\citep{micheli2022transformers}, diffusion-based simulation with DIAMOND~\citep{alonso2024diffusion}, and scalable imagination training with DreamerV4~\citep{hafner2025training}. Recent generative systems further extend world modeling toward interactive environments and large-scale physical simulation, including Oasis~\citep{decart2024oasis}, Genie~\citep{bruce2024genie}, HunyuanWorld~\citep{team2025hunyuanworld}, and Cosmos~\citep{agarwal2025cosmos}.

\noindent\textbf{Joint-embedding predictive architectures} (JEPAs)
learn representations by predicting target embeddings from context without requiring observation reconstruction~\citep{lecun2022path}. I-JEPA~\citep{assran2023self} and V-JEPA~\citep{bardes2023v} apply this principle to masked image and video representations. Action-conditioned world models extend representation prediction to the consequences of actions: DINO-WM~\citep{zhou2024dino} predicts future pretrained visual features, V-JEPA 2-AC~\citep{assran2025v} learns action-conditioned dynamics through post-training, and PLDM~\citep{sobal2026learning} jointly learns representations and dynamics from reward-free offline trajectories. LeWM~\citep{maes2026leworldmodel} adopts Sketched Isotropic Gaussian Regularization (SIGReg) from LeJEPA~\citep{balestriero2025lejepa} to prevent representation collapse, enabling stable end-to-end training of the encoder and predictor.
Recent work further studies how model architecture, training objectives, and planner choices affect physical planning performance~\citep{terver2025drives}, and connects JEPA pretraining error to downstream planning regret~\citep{cui2026generalization}.

\noindent\textbf{Latent transition structure and geometry}
characterize how learned representations organize dynamics for prediction and control. E2C~\citep{watter2015embed} and PCC~\citep{levine2019prediction} learn representations suited to locally linear control, while deep Koopman models~\citep{lusch2018deep} seek coordinates with linear evolution. More recent theory examines when predictive representations recover underlying world structure: conditions for recovering latent world variables are characterized in LeJEPA~\citep{klindt2026does}, while controlled transition identifiability is linked to conditional action variation~\citep{zhang2026identifiability}. Complementary work encourages action-sensitive geometry by decoding actions from latent displacements so that different actions induce distinguishable latent changes~\citep{zhang2026delta}. At the trajectory level, temporal straightening~\citep{wang2026temporal} regularizes directional changes between consecutive latent displacements, encouraging straighter representations for planning.

\section{A Transition-Field Formulation and Curvature Intervention}
\label{sec:3}

As shown in Figure~\ref{fig:main}, LeWM jointly trains an encoder \(\mathcal E\) and an action-conditioned predictor \(\mathcal P\), using SIGReg to prevent representation collapse. Given an observation \(o_t\), the encoder produces the current latent state \(z_t\), and the predictor estimates the next latent state \(\hat z_{t+1}\) from \(z_t\) and \(a_t\):
\begin{equation}
z_t=\mathcal E(o_t),
\qquad
\hat z_{t+1}=\mathcal P(z_t,a_t),
\qquad
\mathcal L_{\mathrm{LeWM}}
=
\mathbb E
\left\lVert
\hat z_{t+1}-z_{t+1}
\right\rVert^2
+
\lambda_{\mathrm{sig}}\mathcal L_{\mathrm{sig}}.
\label{eq:lewm_objective}
\end{equation}

\subsection{Transition-Field Formulation}

\noindent\textbf{Latent prediction as a transition field.}
Rather than analyzing the predicted next state directly, we write the prediction as a latent change. Let \(v_t=z_{t+1}-z_t\) denote the encoded displacement and define
\begin{equation}
F(z_t,a_t)
=
\hat z_{t+1}-z_t
=
\mathcal P(z_t,a_t)-z_t.
\label{eq:field_definitions}
\end{equation}
The LeWM objective can then be written as
\begin{equation}
\mathcal L_{\mathrm{LeWM}}
=
\underbrace{
\mathbb E
\left\lVert
F(z_t,a_t)-v_t
\right\rVert^2
}_{\mathcal L_{\mathrm{pred}}}
+
\lambda_{\mathrm{sig}}\mathcal L_{\mathrm{sig}}.
\label{eq:field_formulation}
\end{equation}
Thus, next-state prediction is equivalently the learning of an action-conditioned latent displacement field, which we call the \emph{transition field}.

\noindent\textbf{From a single action to future differences.}
For planning, the key question is how differently this field responds to different actions. We describe its local geometry using the action and state Jacobians:
\begin{equation}
J_aF_t
=
\frac{\partial F(z_t,a_t)}{\partial a_t},
\qquad
J_zF_t
=
\frac{\partial F(z_t,a_t)}{\partial z_t}.
\label{eq:field_jacobians}
\end{equation}
At a fixed latent state, \(J_aF_t\) describes how an action changes the immediate predicted transition. Planning, however, depends on where those action-induced differences lead after several rollout steps.
Starting from \(\hat z_t=z_t\), the learned rollout follows
\begin{equation}
\hat z_{j+1}
=
\hat z_j+F(\hat z_j,a_j).
\label{eq:field_rollout}
\end{equation}
Holding later actions fixed, the effect of \(a_t\) on the predicted state at horizon \(H\) is
\begin{equation}
\frac{\partial \hat z_{t+H}}{\partial a_t}
=
\bigl(I+J_zF_{t+H-1}\bigr)
\cdots
\bigl(I+J_zF_{t+1}\bigr)
J_aF_t,
\label{eq:multistep_action_propagation}
\end{equation}
where \(F_j=F(\hat z_j,a_j)\), and the Jacobians are evaluated along the predicted rollout. The action Jacobian \(J_aF_t\) determines the initial action-induced difference, while the subsequent state Jacobians determine how that difference is transformed over the rollout.

\noindent\textbf{Complete action sequences for planning.}
Planning compares complete candidate action sequences. We therefore extend the single-action propagation above by collecting the effects of all actions on the terminal predicted state. For the stacked sequence
\(\mathbf a=(a_t,\ldots,a_{t+H-1})\),
define
\begin{equation}
\mathcal J_H
=
\frac{\partial \hat z_{t+H}}{\partial \mathbf a}
=
\left[
\frac{\partial \hat z_{t+H}}{\partial a_t},
\ldots,
\frac{\partial \hat z_{t+H}}{\partial a_{t+H-1}}
\right].
\label{eq:rollout_jacobian}
\end{equation}
With the current latent state fixed, consider a small perturbation \(\delta\mathbf a\) to a candidate sequence \(\mathbf a\). The perturbed sequence \(\mathbf a+\delta\mathbf a\) induces a change in the terminal predicted state
\begin{equation}
\delta\hat z_{t+H}
\approx
\mathcal J_H\,\delta\mathbf a.
\label{eq:terminal_action_variation}
\end{equation}
Thus, \(\mathcal J_H\) connects the field's local action sensitivity to the differences among candidate futures at the planning horizon.

\noindent\textbf{Shaping transition geometry through prediction targets.}
To understand how the geometry of the encoded displacement targets can influence the learned field, we examine its variation along an encoded trajectory. Writing \(F_t=F(z_t,a_t)\) and \(\Delta a_t=a_{t+1}-a_t\), a first-order expansion gives
\begin{equation}
F_{t+1}-F_t
\approx
J_zF_t\,v_t+J_aF_t\,\Delta a_t.
\label{eq:field_temporal_geometry}
\end{equation}
The first term captures field variation due to changes in the latent state, while the second captures variation due to changes in the action. Since the field is trained to match encoded displacements, shaping the temporal geometry of these targets provides a way to intervene on the geometry learned by the field. 
Appendix~\ref{app:transition_derivation} relates field variation to local curvature, and Appendix~\ref{app:temporal_geometry} shows that small relative prediction errors imply close agreement between field and target straightness.
\begin{table}[t]
\centering
\small
\caption{Closed-loop CEM success rates (\%) across curvature regularization strengths. Results are the mean \(\scriptstyle\pm\) standard deviation over three independent training seeds. Green entries highlight representative high-performing settings within each task.}
\label{tab:curvature_lambda_sweep}
\begin{tabular}{rcccc}
\toprule
\(\lambda\) & TwoRoom & PushT & Reacher & Cube \\
\midrule
\(0\) (Base)
& \(90.0 \scriptstyle\pm 1.0\)
& \(94.7 \scriptstyle\pm 1.5\)
& \(82.3 \scriptstyle\pm 2.9\)
& \(76.7 \scriptstyle\pm 4.2\) \\
\midrule
\(0.0005\)
& \(89.3 \scriptstyle\pm 1.5\)
& \(93.7 \scriptstyle\pm 2.1\)
& \(81.7 \scriptstyle\pm 5.0\)
& \(75.3 \scriptstyle\pm 2.1\) \\
\(0.001\)
& \(90.0 \scriptstyle\pm 2.0\)
& \cellcolor{cellgreen}\color{fontgreen}{\(95.3 \scriptstyle\pm 2.1\)}
& \(82.7 \scriptstyle\pm 2.5\)
& \(77.0 \scriptstyle\pm 1.7\) \\
\(0.002\)
& \(91.7 \scriptstyle\pm 2.1\)
& \cellcolor{cellgreen}\color{fontgreen}{\(94.7 \scriptstyle\pm 2.5\)}
& \(84.0 \scriptstyle\pm 3.0\)
& \(74.0 \scriptstyle\pm 3.5\) \\
\(0.005\)
& \(90.7 \scriptstyle\pm 1.5\)
& \(93.3 \scriptstyle\pm 0.6\)
& \cellcolor{cellgreen}\color{fontgreen}{\(88.0 \scriptstyle\pm 3.0\)}
& \(77.3 \scriptstyle\pm 2.1\) \\
\(0.01\)
& \(92.3 \scriptstyle\pm 1.2\)
& \(92.7 \scriptstyle\pm 1.5\)
& \(86.0 \scriptstyle\pm 2.0\)
& \(77.3 \scriptstyle\pm 2.5\) \\
\(0.02\)
& \(94.0 \scriptstyle\pm 2.0\)
& \(87.3 \scriptstyle\pm 2.5\)
& \(80.3 \scriptstyle\pm 4.6\)
& \(78.3 \scriptstyle\pm 2.9\) \\
\(0.05\)
& \(94.0 \scriptstyle\pm 1.0\)
& \(75.3 \scriptstyle\pm 6.0\)
& \(82.7 \scriptstyle\pm 3.5\)
& \(79.3 \scriptstyle\pm 1.5\) \\
\(0.1\)
& \(96.7 \scriptstyle\pm 0.6\)
& \(54.7 \scriptstyle\pm 6.7\)
& \(81.0 \scriptstyle\pm 5.2\)
& \(85.0 \scriptstyle\pm 4.6\) \\
\(0.2\)
& \cellcolor{cellgreen}\color{fontgreen}{\(99.0 \scriptstyle\pm 0.0\)}
& \(36.3 \scriptstyle\pm 1.5\)
& \(82.0 \scriptstyle\pm 2.6\)
& \cellcolor{cellgreen}\color{fontgreen}{\(86.3 \scriptstyle\pm 2.9\)} \\
\(0.5\)
& \cellcolor{cellgreen}\color{fontgreen}{\(99.0 \scriptstyle\pm 0.0\)}
& \(20.7 \scriptstyle\pm 3.1\)
& \(84.3 \scriptstyle\pm 2.5\)
& \cellcolor{cellgreen}\color{fontgreen}{\(87.3 \scriptstyle\pm 0.6\)} \\
\bottomrule
\end{tabular}
\end{table}

\subsection{Curvature Intervention}
To intervene on the learned field through its prediction targets, we use curvature regularization~\citep{wang2026temporal}, which penalizes directional changes between consecutive encoded displacements while keeping the model architecture and planner fixed.
\begin{equation}
\mathcal L_{\mathrm{curv}}
=
\mathbb E\left[1-\cos(v_t,v_{t+1})\right],
\qquad
\mathcal L
=
\mathcal L_{\mathrm{LeWM}}
+
\lambda\mathcal L_{\mathrm{curv}}.
\label{eq:straightening_objective}
\end{equation}

A larger \(\lambda\) therefore encourages consecutive latent displacements to point in more similar directions. By varying \(\lambda\), we can directly test how changes in this geometry affect both the learned transition field and planning.
We train three models with independent seeds per task and $\lambda$. 
On held-out trajectories across all four tasks, the base transition fields closely reproduce the encoded displacements and their mean temporal straightness on held-out trajectories, as shown in Table~\ref{tab:field_temporal_geometry}.

\noindent\textbf{Task-dependent planning regimes.}
We sweep the intervention strength \(\lambda\) and report closed-loop CEM success rates in Table~\ref{tab:curvature_lambda_sweep}.
For subsequent analyses, we use \(\lambda=0.5\), \(0.001\), \(0.005\), and \(0.5\) as the beneficial regimes for TwoRoom, PushT, Reacher, and Cube, respectively, and examine \(\lambda=0.1\) as an over-regularized regime for PushT.

TwoRoom and Cube achieve their highest mean success toward the upper end of the sweep.
Reacher achieves its highest mean success at \(\lambda=0.005\).
PushT stays close to its base performance under weak regularization and deteriorates rapidly at larger values of \(\lambda\).
We use these regimes to examine how beneficial and excessive regularization reshape the learned transition field.

\section{How Does Curvature Reshape the Field?}
\label{sec:4}
Planning compares candidate action sequences starting from the same current state, and Equation~\ref{eq:multistep_action_propagation} shows how action-induced differences are propagated by subsequent state dynamics. We therefore begin by examining how curvature changes action sensitivity within a single prediction step.

\begin{figure}[t]
\centering
\includegraphics[width=\textwidth]{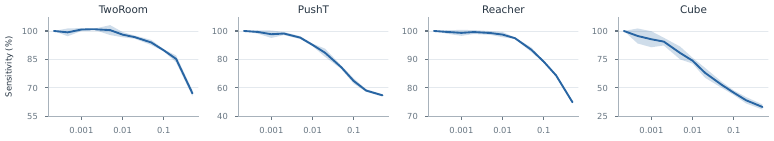}
\caption{One-step action sensitivity across curvature regularization strengths, normalized to the base-model level of 100\%. Curves show means, with shaded bands indicating standard deviations.}
\label{fig:action_contraction_sweep}
\end{figure}

\noindent\textbf{Curvature reduces one-step action sensitivity.}
At a fixed latent state, small changes in the action input can
alter the one-step predicted transition. We measure total action
sensitivity by
\begin{equation}
S_a=\mathbb E\left[\lVert J_aF\rVert\right].
\label{eq:action_geometry_metrics}
\end{equation}
Figure~\ref{fig:action_contraction_sweep} reports one-step action sensitivity relative to the base model within each task, showing that sensitivity generally decreases with stronger curvature regularization across all four tasks, with small local non-monotonicities.

\noindent\textbf{The non-affine component carries most of the contraction.}
To identify which part of the learned dynamics carries this contraction, we use a global affine model as a reference, whose action Jacobian is constant across states and actions.
Let \(z_t\in\mathbb R^d\) and \(a_t\in\mathbb R^q\). For each task and checkpoint, we fit
\begin{equation}
L(z_t,a_t)
=
Az_t+Ba_t+b,
\qquad
A\in\mathbb R^{d\times d},
\quad
B\in\mathbb R^{d\times q},
\quad
b\in\mathbb R^d,
\label{eq:linear_field_fit}
\end{equation}
using all valid transitions from 400 fitting trajectories and evaluate it on 100 held-out trajectories. 
Appendix~\ref{app:affine_reference} shows that the affine reference explains 12.7–63.1\% of the held-out transition-field variation across checkpoints in the base and beneficial regimes. We retain the remaining structure as the non-affine component,
\begin{equation}
G(z_t,a_t)=F(z_t,a_t)-L(z_t,a_t),
\qquad
J_aF=B+J_aG.
\label{eq:linear_field_decomposition}
\end{equation}
Here, \(J_aG\) captures the local action response left unexplained by the affine reference.
To quantify how these components contribute to the contraction, we decompose the full one-step action-response energy as
\begin{equation}
E_F
=
\mathbb E\left[\lVert J_aF\rVert^2\right]
=
\underbrace{\lVert B\rVert^2}_{E_L}
+
\underbrace{\mathbb E\left[\lVert J_aG\rVert^2\right]}_{E_G}
+
\underbrace{2\,\mathbb E\left[\langle B,J_aG\rangle\right]}_{E_X}.
\label{eq:app_action_energy_decomposition}
\end{equation}
Here, \(E_L\), \(E_G\), and \(E_X\) are the affine, non-affine, and cross-term contributions, respectively. Norms and inner products are Frobenius.

\begin{table}[ht]
\centering
\small
\caption{Contributions to one-step action-response energy reduction.
For each training seed, reductions are expressed as percentages of the corresponding base model's total energy. Positive values indicate reductions, and negative values offset the reduction.}
\label{tab:app_action_energy_components}
\begin{tabular}{lcc>{\columncolor[gray]{0.93}}ccc}
\toprule
& TwoRoom & \multicolumn{2}{c}{PushT} & Reacher & Cube \\
\cmidrule(lr){3-4}
\(\lambda\)
& \(0.5\) & \(0.001\) & \(0.1\) & \(0.005\) & \(0.5\) \\
\midrule
Affine \(E_L\)
& \(-0.13{\scriptstyle\pm0.05}\)
& \(0.04{\scriptstyle\pm0.04}\)
& \(1.35{\scriptstyle\pm0.05}\)
& \(0.00{\scriptstyle\pm0.01}\)
& \(3.73{\scriptstyle\pm0.20}\) \\
Non-affine \(E_G\)
& \(39.28{\scriptstyle\pm2.53}\)
& \(4.19{\scriptstyle\pm3.33}\)
& \(51.40{\scriptstyle\pm1.66}\)
& \(1.36{\scriptstyle\pm0.96}\)
& \(85.67{\scriptstyle\pm1.27}\) \\
Cross term \(E_X\)
& \(0.11{\scriptstyle\pm0.09}\)
& \(-0.01{\scriptstyle\pm0.04}\)
& \(-0.93{\scriptstyle\pm0.11}\)
& \(0.00{\scriptstyle\pm0.02}\)
& \(-6.05{\scriptstyle\pm0.41}\) \\
\midrule
Total reduction
& \(39.26{\scriptstyle\pm2.50}\)
& \(4.22{\scriptstyle\pm3.40}\)
& \(51.81{\scriptstyle\pm1.66}\)
& \(1.36{\scriptstyle\pm0.96}\)
& \(83.35{\scriptstyle\pm1.07}\) \\
\bottomrule
\end{tabular}
\end{table}

Table~\ref{tab:app_action_energy_components} reports how each energy term changes from the base to the regularized checkpoint.
This attribution shows that the net reduction in one-step action-response energy is concentrated in the non-affine component \(E_G\).

\section{How Does the Reshaped Field Affect the Planner?}
\label{sec:5}
To examine how the non-affine component affects planning, we directly scale its action-induced differences and measure the effects on CEM candidate selection and closed-loop performance.

\noindent\textbf{Directly intervening on non-affine action differences.}
For each frozen checkpoint, we construct an intervention that preserves the predicted transition at the CEM proposal center while selectively scaling how the non-affine component distinguishes candidate actions. For a fixed latent state \(z\) and proposal mean action \(\bar a\),
\begin{equation}
F(z,a)
=
F(z,\bar a)
+
B(a-\bar a)
+
\bigl[G(z,a)-G(z,\bar a)\bigr],
\end{equation}
where \(F=L+G\) and \(L(z,a)=Az+Ba+b\). We define
\begin{equation}
F_{\beta}(z,a)
=
F(z,\bar a)
+
B(a-\bar a)
+
\beta\bigl[G(z,a)-G(z,\bar a)\bigr].
\label{eq:beta_field}
\end{equation}
Thus, \(\beta=1\) recovers the original field, \(0\leq\beta<1\) reduces the non-affine differences between candidate actions, and \(\beta>1\) amplifies them. At \(\beta=0\), the prediction at the proposal center and the affine action difference are preserved, while the non-affine difference between candidate actions is removed. Holding \(z\) and \(\bar a\) fixed,
\begin{equation}
J_aF_{\beta}(z,a)
=
B+\beta J_aG(z,a),
\label{eq:beta_jacobian}
\end{equation}
so scaling these non-affine action differences also scales their local action sensitivity.

We first test how this rescaling affects the quality of elite selection.
For each evaluation start, we draw a set \(\mathcal A\) of 300 action sequences from the initial CEM proposal and reuse it at every \(\beta\).
Executing a sequence \(\mathbf a\) in the simulator gives its terminal distance to the goal \(d(\mathbf a)\).
The 30 sequences that finish closest to the goal form \(\mathcal A^\star\), and the distances across all 300 sequences range from \(d_{\min}\) to \(d_{\max}\).
At each \(\beta\), we select the 30 sequences with the lowest predicted costs as the model's elites \(\mathcal A_\beta\).
We compare the two sets using normalized elite regret,
\begin{equation}
\label{eq:regret}
r_\beta =
\frac{\bar d(\mathcal A_\beta)-\bar d(\mathcal A^\star)}
     {d_{\max}-d_{\min}},
\end{equation}
where \(\bar d\) is the mean simulator distance over the indicated set.
When all distances are equal, we assign zero regret.
We report the mean regret over all 50 evaluation starts for each \(\beta\).

\noindent\textbf{Non-affine action differences shape candidate selection and planning.}
We next run closed-loop CEM planning at
\(\beta\in\{0,0.25,0.5,0.75,1,1.25,1.5\}\)
for the base and beneficial checkpoints of each task, as well as the over-regularized PushT checkpoint at \(\lambda=0.1\).

\begin{figure}[t]
\centering
\includegraphics[width=\textwidth]{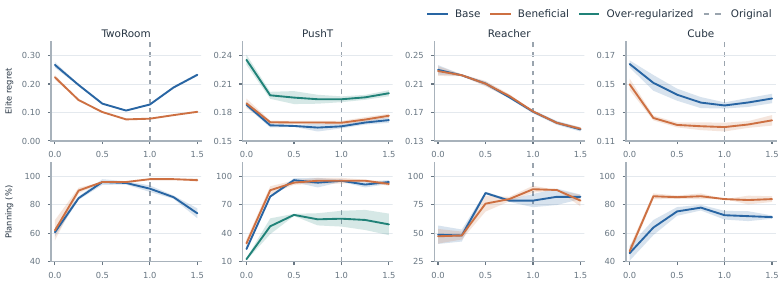}
\caption{Effect of scaling non-affine action differences on elite selection and closed-loop success.
The top row shows normalized elite regret for the initial CEM proposal, with lower values indicating better elite selection.
The bottom row shows closed-loop success.
Curves show means, with shaded bands indicating standard deviations.
The dashed line marks the original scale \(\beta=1\).}
\label{fig:beta_selection_planning}
\end{figure}

Figure~\ref{fig:beta_selection_planning} shows that rescaling non-affine action differences in frozen checkpoints affects both initial CEM elite selection and closed-loop success.
Removing these differences entirely at \(\beta=0\) increases mean elite regret and lowers mean success relative to the original scale in all nine checkpoint settings.
At nonzero scales, elite regret is lowest under moderate attenuation in TwoRoom and decreases as \(\beta\) increases in Reacher.
The effect on closed-loop success can differ even within a task.
In Reacher, reducing \(\beta\) from \(1\) to \(0.5\) increases elite regret in both regimes, raises mean success in the base regime, and lowers it in the beneficial regime.
As rescaling also changes prediction error, Appendix~\ref{app:beta_prediction_error} compares settings with similar mean five-step prediction errors and finds that differences in closed-loop success persist.

\section{How Does Transition Geometry Enter Planning?}
\label{sec:6}

To better understand these planning effects, we next trace how one-step action responses propagate to the planning horizon, where the planner compares predicted goal distances.

\subsection{Action Responses at the Planning Horizon}
\noindent\textbf{Measuring rollout sensitivity.}
For each checkpoint, the rollout Jacobian \(\mathcal J_H\) from Equation~\ref{eq:rollout_jacobian} describes how the predicted terminal latent state responds to small changes in the candidate action sequence. Let \(\Sigma_{\mathbf a}\) denote the covariance of the action-sequence perturbations. To account for the size of these action changes, we use
\begin{equation}
\widetilde{\mathcal J}_H
=
\mathcal J_H\Sigma_{\mathbf a}^{1/2}.
\label{eq:whitened_terminal_action_jacobian}
\end{equation}
We use the same zero-action reference sequence and \(\Sigma_{\mathbf a}=I\) across checkpoints, so all action coordinates are evaluated at the same perturbation scale.

The overall size of this response does not tell us which directions are most affected. Since the planner compares candidates by their predicted distance to the goal, we next separate changes along the goal direction from changes that shift the prediction sideways.

Let \(\hat z_{t+H}\) be the predicted terminal latent state at this reference action sequence, and \(z_g=\mathcal E(o_g)\) the encoded goal. The unit direction from the predicted terminal state to the goal is
\begin{equation}
u
=
\frac{z_g-\hat z_{t+H}}
     {\lVert z_g-\hat z_{t+H}\rVert}.
\label{eq:terminal_goal_direction}
\end{equation}
We measure the total response energy and its contributions along \(u\) and in the remaining directions, orthogonal to \(u\):
\begin{equation}
S_{\mathrm{total}}
=\lVert\widetilde{\mathcal J}_H\rVert^2,\qquad
S_{\parallel}
=\lVert\widetilde{\mathcal J}_H^\top u\rVert^2,\qquad
S_{\perp}
=S_{\mathrm{total}}-S_{\parallel}.
\label{eq:goal_response_energy}
\end{equation}
Here, energy measures how much the predicted terminal state changes when we slightly change the action sequence. We quantify this using the average squared change in latent space. \(S_{\mathrm{total}}\) measures the overall change, \(S_{\parallel}\) measures the change along the goal direction, and \(S_{\perp}\) measures the change in the remaining directions.

\noindent\textbf{Overall contraction and relative goal-direction retention.}
We compare each regularized checkpoint with the base model using
\begin{equation}
R_{\mathrm{total}}
=
\frac{S_{\mathrm{total}}}
     {S_{\mathrm{total}}^{\mathrm{base}}},
\qquad
q
=
\frac{S_{\parallel}/S_{\perp}}
     {S_{\parallel}^{\mathrm{base}}/S_{\perp}^{\mathrm{base}}}.
\label{eq:rollout_response_ratios}
\end{equation}
The ratios use energies averaged over the same held-out states, with superscript \(\mathrm{base}\) denoting the base model.
\(R_{\mathrm{total}}\) records the fraction of total response energy retained relative to the base model. \(R_{\mathrm{total}}<1\) means that nearby candidate action sequences produce more similar predicted futures overall.
\(q\) measures how the ratio of response energy along the goal direction to energy in the remaining directions changes relative to the base model. \(q>1\) means that, compared with the base model, a larger share of the differences between predicted futures lies along the goal direction.
\begin{figure}[t]
\centering
\includegraphics[width=\linewidth]{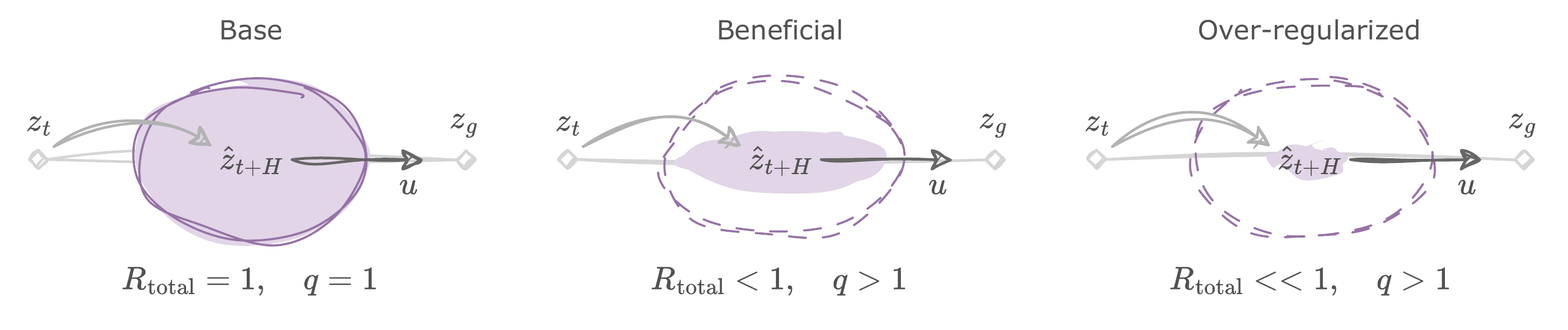}
\caption{Illustration of rollout-response contraction.
Moderate contraction makes goal-direction differences relatively more prominent, while excessive contraction weakens candidate distinctions.
Shaded regions show terminal-state responses to matched action perturbations; dashed contours indicate the base response.}
\label{fig:curv_shape}
\end{figure}
\begin{figure}[t]
\centering
\includegraphics[width=\textwidth]{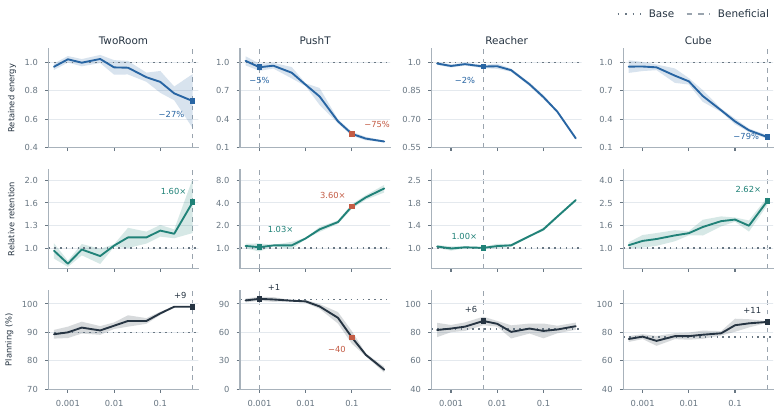}
\caption{Rollout geometry and closed-loop CEM success across curvature regularization strengths.
\(R_{\mathrm{total}}<1\) indicates overall contraction, while \(q>1\) indicates greater relative retention of response energy along the goal direction.
Curves show means, with shaded bands indicating standard deviations.
Vertical dashed lines mark the beneficial regimes.}
\label{fig:propagation_lambda}
\end{figure}

Figure~\ref{fig:curv_shape} illustrates how these two quantities describe changes in the rollout response. Appendix~\ref{app:rollout_geometry} provides the covariance interpretation, connection to the goal cost, and the linear case.

\noindent\textbf{Rollout responses and planning across tasks.}
Figure~\ref{fig:propagation_lambda} shows that stronger regularization generally reduces \(R_{\mathrm{total}}\) and increases \(q\) across the four tasks.

This pattern suggests a geometric explanation for the planning gains. As the overall response contracts, differences that change predicted goal distance become relatively more prominent. When the goal-direction response remains large enough, small changes in candidate actions can still produce distinct predicted goal distances while causing less variation in other directions. Excessive contraction, however, may also weaken these goal-distance differences, making candidate actions harder to distinguish despite their larger relative share.

The relationship between these geometric changes and planning performance differs across tasks. In TwoRoom and Cube, substantial gains accompany overall contraction and greater relative retention along the goal direction. In Reacher, the more modest planning gain occurs with both geometry measures close to their base values. PushT already achieves strong base performance, and weak regularization leaves both planning success and rollout geometry close to their base levels. At \(\lambda=0.1\), both the total response and its goal-direction component contract as planning deteriorates, even as \(q\) rises. This suggests that the absolute separation along the goal direction matters even when it accounts for a larger share of the remaining response.

\noindent\textbf{Generalization of rollout geometry.}
For a fixed reference action sequence and perturbation covariance, both diagnostics are defined in terms of the rollout Jacobian and the latent goal direction so that the same analysis can be applied across solvers and differentiable latent world models. 
On LeWM, the main planning trends are broadly consistent across five sampling- and gradient-based solvers.
On DINO-WM, we observe overall contraction and greater relative goal-direction retention on all four tasks.
Appendix~\ref{app:rollout_generalization} reports both evaluations.
\begin{figure}[t]
\centering
\includegraphics[width=\linewidth]{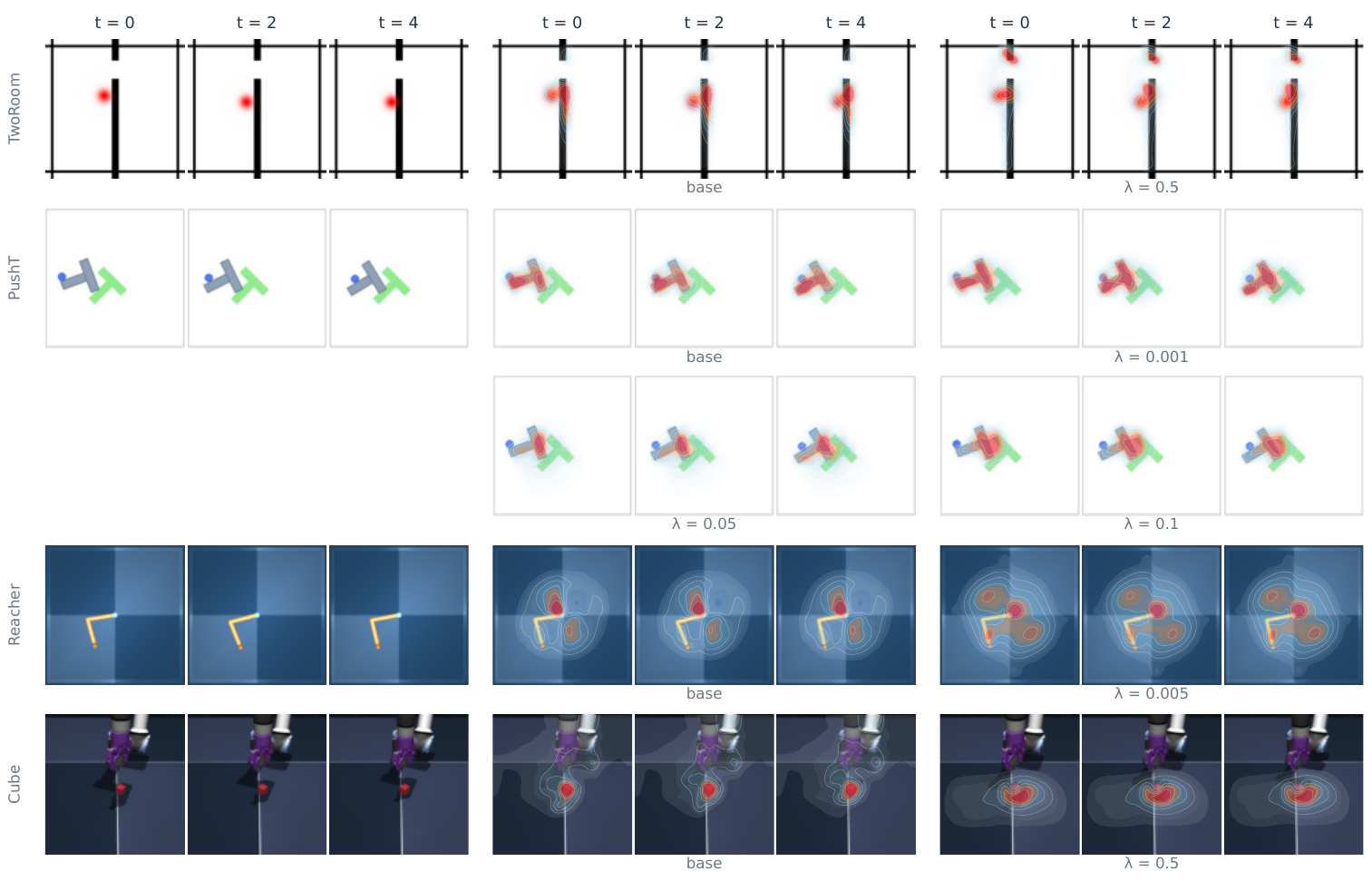}
\caption{Encoder attention across curvature regularization strengths. Maps compare the base and beneficial models on the same observations, with additional stronger regularization settings for PushT. Colors are normalized per panel and indicate relative spatial emphasis.}
\label{fig:attention_maps}
\end{figure}

\subsection{Encoder Attention}
\label{sec:attention}
We now return to the observations that LeWM encodes as latent states \(E(o_t)\).
As Equation~\ref{eq:multistep_action_propagation} shows, the encoded state conditions the response to the first action and initializes the rollout through which that response propagates.
We therefore compare encoder attention maps on the same observations across regularization strengths to examine whether changes in rollout geometry are accompanied by shifts in the visual structure emphasized by the encoder.

The examples in Figure~\ref{fig:attention_maps} show different shifts in attention across tasks.
In TwoRoom, attention becomes more focused on the agent and the passage through which it moves.
In PushT, the beneficial model emphasizes both the T-shaped object and the pusher.
At \(\lambda=0.1\), attention shifts away from both toward the image center, accompanying the sharp rollout contraction and drop in planning success shown in Figure~\ref{fig:propagation_lambda}.
In Reacher, attention becomes more broadly distributed around the arm, with fewer isolated hotspots.
In Cube, attention shifts toward the cube and the surrounding tabletop, with less emphasis on the robot.
More examples are provided in Appendix~\ref{app:attention_map}.

The rollout results suggest that both the magnitude of action-induced differences and their alignment with the goal direction matter for planning.

\section{Discussion}
We study how latent transition geometry shapes planning through an action-conditioned transition-field view of world models.
Using curvature regularization as a controlled intervention, we first ask how it changes the one-step action response. We find that most of the reduction in action-response energy comes from the non-affine component.
To assess its role in planning, we rescale the action-induced differences in this component in frozen models. This changes the quality of the elites selected by CEM and affects closed-loop success, with effects that vary across tasks and checkpoints.
At the planning horizon, stronger curvature regularization generally reduces the overall response to actions, while differences along the goal direction become relatively more prominent.
When differences in predicted goal distance remain large enough, the planner can still distinguish candidates even as their predicted futures become more compact.
Excessive contraction can weaken these goal-distance differences, making candidates harder to distinguish and potentially reducing closed-loop success.
This balance helps explain why the planning effects of regularization vary across tasks.
More broadly, the transition-field view complements one-step prediction metrics and offers a potentially useful framework for understanding planning failures.

Our experiments cover four continuous-control tasks, with broadly consistent planning trends across five sampling- and gradient-based solvers on LeWM. 
We also observe similar rollout-geometry trends on DINO-WM across the same four tasks.
Qualitative attention maps show accompanying shifts in visual emphasis across regularization regimes.
The Jacobian measurements describe responses to small action changes around observed inputs and fixed reference action sequences, so they cover only part of the geometry encountered during planning.
Evaluations on additional world models and at longer horizons would help establish how broadly these findings apply.

\newpage
\subsubsection*{Reproducibility Statement}
This work presents an empirical analysis of learned latent dynamics, with the curvature intervention and analysis procedures defined in the paper. Appendix~\ref{app:experimental_settings} provides detailed training settings, checkpoint selection, planner configuration, and analysis protocols.


\subsubsection*{Acknowledgments}
The authors acknowledge the use of resources provided by the Isambard-AI~\citep{mcintosh2024isambard} National AI Research Resource (AIRR), operated by the University of Bristol and funded by the UK Government's Department for Science, Innovation and Technology (DSIT) via UK Research and Innovation and the Science and Technology Facilities Council [ST/AIRR/I-A-I/1023].

\bibliography{iclr2027_conference}
\bibliographystyle{iclr2027_conference}

\appendix
\newpage
\newpage
\section{Experimental Setup}
\label{app:experimental_settings}

\noindent\textbf{LeWM training and CEM planning.}
We follow the LeWM training setup~\citep{maes2026leworldmodel}.
Within each task, we keep the architecture and training configuration fixed and vary only the curvature regularization weight \(\lambda\).
All LeWM analyses use the full latent and action history, which we omit from the main-text notation for brevity.
Table~\ref{tab:experimental_settings} summarizes the LeWM training and CEM settings used in the main experiments.

\begin{table}[ht]
\centering
\caption{Model, training, and planning settings for the LeWM experiments with CEM.}
\label{tab:experimental_settings}
\begin{tabular}{ll}
\toprule
Setting & Value \\
\midrule
\multicolumn{2}{l}{\textit{Context}} \\
History length & 3 \\
Prediction steps & 1 \\
Frameskip & 5 \\
\midrule
\multicolumn{2}{l}{\textit{Training}} \\
Batch size & 128 \\
Optimizer & AdamW \\
Learning rate & \(5\times10^{-5}\) \\
Training epochs & 10 \\
Training precision & BF16 \\
SIGReg weight & 0.09 \\
\midrule
\multicolumn{2}{l}{\textit{CEM planning}} \\
Planning horizon \(H\) & 5 \\
Candidates per iteration & 300 \\
Elites per iteration & 30 \\
Optimization iterations & 30 \\
Evaluation episodes & 50 \\
\bottomrule
\end{tabular}
\end{table}

\noindent\textbf{Main-text evaluation protocols.}
We aggregate the main quantitative results across three training seeds, computing normalized diagnostics within each seed before aggregation.
For the CEM curvature sweep, we average the success rates from evaluation seeds 42 and 43 within each checkpoint before reporting the mean and sample standard deviation across training seeds.
The \(\beta\)-intervention uses the same 50 evaluation starts from seed 42 across checkpoints and response scales within each task.
Field--target agreement, one-step action sensitivity, and non-affine attribution are evaluated on the same 100 held-out trajectories per task.
For rollout geometry, we use 50 matched held-out state--goal pairs per task.
The attention visualizations compare checkpoints on the same observations.

\noindent\textbf{LeWM planning with additional solvers.}
We extend the LeWM planning evaluation to four additional solvers within the \texttt{stable-worldmodel} framework~\citep{maes2026stable}.
All five solvers share one checkpoint per task and regularization strength and use the same 50 episodes sampled with evaluation seed 42 within each task.
For the sampling-based solvers, iCEM and MPPI, like CEM, run for 30 iterations with 300 candidates and top-30 selection per iteration.
For the gradient-based solvers, multi-start projected GD optimizes 32 initial action sequences for 25 steps each, using a learning rate of \(0.01\) on PushT and \(0.001\) on the other tasks.
Adam uses 100 initial action sequences with 30 steps per sequence.

\noindent\textbf{DINO-WM training and evaluation.}
Following the temporal straightening setup~\citep{wang2026temporal}, we train DINO-WM on TwoRoom, PushT, Reacher, and Cube at \(\lambda\in\{0,0.01,0.1\}\) for two epochs with a frozen DINOv2 backbone.
We use Adam at \(1\times10^{-6}\) for the global projector and \(3\times10^{-4}\) for the decoder, and AdamW at \(5\times10^{-4}\) for the predictor, action encoder, and proprioception encoder.
We use training seed 0 and evaluate each checkpoint with CEM over 50 episodes.
On PushT, we follow the original DINO-WM evaluation protocol with dataset goals.
For the other tasks, we use the \texttt{stable-worldmodel} environments and sample held-out start--goal pairs with evaluation seed 42.

\section{Transition-Field Analysis}

\subsection{Action Propagation and Temporal Curvature}
\label{app:transition_derivation}

\noindent\textbf{Propagation of action effects.}
For a predictor with history, the rollout update and readout are
\[
h_{j+1}=T(h_j,a_j),
\qquad
\hat z_j=Rh_j.
\]
Holding the initial history and all other actions fixed, the chain rule gives
\begin{equation}
\frac{\partial \hat z_{t+H}}{\partial a_{t+k}}
=
R\,J_hT_{t+H-1}\cdots J_hT_{t+k+1}\,J_aT_{t+k},
\qquad k=0,\ldots,H-1.
\label{eq:app_multistep_action_propagation}
\end{equation}
The Jacobians are evaluated along the predicted rollout, with an empty product equal to the identity.
These blocks form \(\mathcal J_H\). Taking the state itself as the history recovers Equation~\ref{eq:multistep_action_propagation} for the first action.

\noindent\textbf{Local temporal curvature.}
Along an encoded trajectory, write
\[
F_t=F(z_t,a_t),
\qquad
v_t=z_{t+1}-z_t,
\qquad
\Delta a_t=a_{t+1}-a_t.
\]
For \(F_t\ne0\), unit normalization removes the component of the first-order variation parallel to \(F_t\).
Applying this to the field expansion in Equation~\ref{eq:field_temporal_geometry} gives, as \(\eta_t=\lVert(v_t,\Delta a_t)\rVert\to0\),
\begin{equation}
1-\cos(F_t,F_{t+1})
=
\frac{\left\lVert
P_{F_t}^{\perp}\left(J_zF_t\,v_t+J_aF_t\,\Delta a_t\right)
\right\rVert^2}
{2\lVert F_t\rVert^2}
+o(\eta_t^2),
\label{eq:app_curvature_field_relation}
\end{equation}
where \(P_{F_t}^{\perp}\) projects onto the orthogonal complement of \(F_t\).

\subsection{Field--Target Agreement}
\label{app:temporal_geometry}

To assess when the temporal geometry of the displacement targets carries over to the learned field, we relate their straightness scores to the one-step prediction error.
State and displacement prediction have identical errors,
\begin{equation}
\epsilon_t=\mathcal P(z_t,a_t)-z_{t+1}=F_t-v_t.
\label{eq:app_field_identity}
\end{equation}
For nonzero \(x\), write \(n(x)=x/\lVert x\rVert\).
For nonzero \(F_t,F_{t+1},v_t,v_{t+1}\), define
\(s_t^F=\cos(F_t,F_{t+1})\) and \(s_t^v=\cos(v_t,v_{t+1})\).
The triangle and reverse triangle inequalities give, for nonzero \(x,y\),
\[
\lVert n(x)-n(y)\rVert
\le
\frac{\lvert\lVert y\rVert-\lVert x\rVert\rvert+\lVert x-y\rVert}{\lVert y\rVert}
\le
\frac{2\lVert x-y\rVert}{\lVert y\rVert}.
\]
Adding and subtracting \(n(v_t)^\top n(F_{t+1})\) in the straightness difference and applying Cauchy--Schwarz gives
\begin{equation}
\begin{aligned}
\left|s_t^F-s_t^v\right|
&\le \lVert n(F_t)-n(v_t)\rVert+\lVert n(F_{t+1})-n(v_{t+1})\rVert\\
&\le 2\frac{\lVert\epsilon_t\rVert}{\lVert v_t\rVert}
+2\frac{\lVert\epsilon_{t+1}\rVert}{\lVert v_{t+1}\rVert}.
\end{aligned}
\label{eq:app_field_target_straightness}
\end{equation}
Thus, small relative prediction errors imply similar field and target straightness.

We next check how closely the base transition fields reproduce their displacement targets and mean temporal straightness on the same 100 held-out trajectories within each task.
To measure displacement prediction, we compute
\begin{equation}
R^2(F,v)
=
1-
\frac{\sum_{i=1}^{N}\lVert v_i-F_i\rVert^2}
{\sum_{i=1}^{N}\lVert v_i-\bar v\rVert^2},
\end{equation}
where \(i\) indexes the \(N\) held-out transitions and \(\bar v\) is their mean displacement.
For temporal straightness, we compare the averages
\begin{equation}
S_v=\mathbb E[s_t^v],
\qquad
S_F=\mathbb E[s_t^F].
\end{equation}
Table~\ref{tab:field_temporal_geometry} shows that the base transition fields closely reproduce the encoded displacements and match their mean temporal straightness across all four tasks.
This agreement supports using the temporal geometry of the displacement targets to guide interventions on the learned field.

\begin{table}[ht]
\centering
\small
\caption{Agreement between the base transition field and encoded displacements, evaluated on 100 held-out trajectories per task.}
\label{tab:field_temporal_geometry}
\begin{tabular}{rcccc}
\toprule
& TwoRoom & PushT & Reacher & Cube \\
\midrule
\(R^2(F,v)\)
& \(0.995{\scriptstyle\pm0.000}\)
& \(0.982{\scriptstyle\pm0.000}\)
& \(0.997{\scriptstyle\pm0.000}\)
& \(0.992{\scriptstyle\pm0.001}\) \\
\(\lvert S_F-S_v\rvert\)
& \(0.004{\scriptstyle\pm0.001}\)
& \(0.030{\scriptstyle\pm0.002}\)
& \(0.004{\scriptstyle\pm0.000}\)
& \(0.013{\scriptstyle\pm0.001}\) \\
\bottomrule
\end{tabular}
\end{table}

\subsection{Global Affine Reference}
\label{app:affine_reference}

The preceding analysis evaluates how well the learned field reproduces its displacement targets.
For the action-response decomposition in Section~\ref{sec:4}, we fit the global affine reference \(L\) defined in Equation~\ref{eq:linear_field_fit} to each learned field \(F\).
Five-fold cross-validation at the trajectory level determines the ridge penalty.
The final fit uses all valid transitions from 400 trajectories, with another 100 trajectories reserved for evaluation.
The held-out coefficient of determination is
\begin{equation}
R_F^2
=
1-
\frac{\sum_{i=1}^{N}\lVert F_i-L_i\rVert^2}
{\sum_{i=1}^{N}\lVert F_i-\bar F\rVert^2},
\qquad
\bar F=\frac{1}{N}\sum_{i=1}^{N}F_i,
\label{eq:app_linear_r2}
\end{equation}
where \(F_i\) and \(L_i\) are the learned field and affine predictions on the \(i\)-th held-out transition, and \(N\) is the number of such transitions.

Table~\ref{tab:app_linear_fit} reports \(R_F^2\) for the base and beneficial regimes.
The beneficial settings use \(\lambda=0.5\), \(0.001\), \(0.005\), and \(0.5\) for TwoRoom, PushT, Reacher, and Cube, respectively.
Across all four tasks, the global affine reference leaves substantial held-out field variation unexplained in both regimes.
This motivates retaining the residual as the non-affine component in the action-response decomposition.

\begin{table}[ht]
\centering
\small
\caption{Held-out \(R_F^2\) of the global affine reference in the base and beneficial regimes. Entries report means and sample standard deviations across three training seeds.}
\label{tab:app_linear_fit}
\begin{tabular}{lcccc}
\toprule
& TwoRoom & PushT & Reacher & Cube \\
\midrule
Base (\(\lambda=0\))
& \(0.401{\scriptstyle\pm0.001}\)
& \(0.405{\scriptstyle\pm0.005}\)
& \(0.129{\scriptstyle\pm0.001}\)
& \(0.468{\scriptstyle\pm0.005}\) \\
Beneficial
& \(0.328{\scriptstyle\pm0.002}\)
& \(0.409{\scriptstyle\pm0.005}\)
& \(0.127{\scriptstyle\pm0.001}\)
& \(0.631{\scriptstyle\pm0.003}\) \\
\bottomrule
\end{tabular}
\end{table}
\section{Prediction Error and Planning}

\subsection{One-Step Error and Candidate Regret}
\label{app:pred_plan_gap}
To illustrate the distinction between one-step predictive accuracy and planning utility, we evaluate the base models on held-out state--goal pairs.
For each pair, we draw 300 candidate action sequences from the initial CEM proposal with horizon \(H=5\).
We measure one-step prediction error by averaging the squared Euclidean error of the first predicted transition across the candidates.

After rolling out the candidates for five steps, we select the sequence with the lowest predicted cost.
We compute normalized regret as defined by Equation~\ref{eq:regret}, with the two sets each containing a single sequence: the model's selection and the best candidate in the simulator, respectively.
We omit pairs with zero range in simulator costs.

\begin{figure}[ht]
\centering
\includegraphics[width=\linewidth]{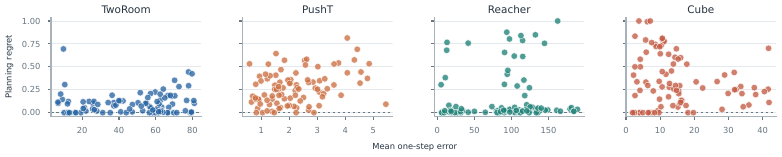}
\caption{One-step prediction error and five-step candidate regret for the base models. Each point represents one held-out state--goal pair evaluated with 300 candidate sequences.}
\label{fig:prediction_planning_gap}
\end{figure}

Figure~\ref{fig:prediction_planning_gap} shows that similar one-step prediction errors can accompany substantially different multi-step planning regret.
This provides a simple empirical illustration of why one-step accuracy alone gives an incomplete picture of planning utility.

\subsection{Prediction Error under Response Scaling}
\label{app:beta_prediction_error}

We next examine whether prediction error at the planning horizon tracks changes in closed-loop success under response scaling.

\noindent\textbf{Five-step error and closed-loop success.}
For each checkpoint, we measure the mean squared Euclidean error between predicted terminal latents and encoded environment observations at five steps, using the same held-out candidate action sequences across \(\beta\).

\begin{figure}[ht]
\centering
\includegraphics[width=\linewidth]{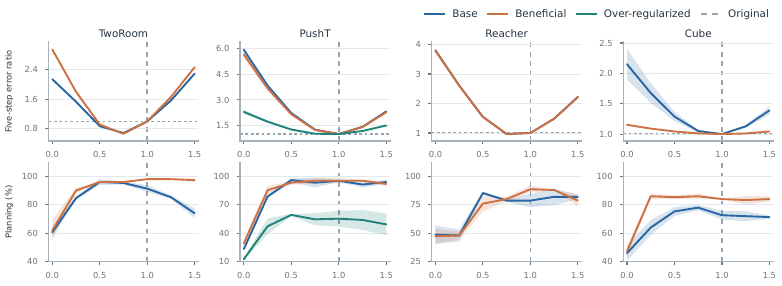}
\caption{Five-step prediction error and closed-loop success under response scaling. 
The top row shows five-step error normalized to its value at \(\beta=1\) for the same checkpoint. 
The bottom row shows closed-loop success. Vertical dashed lines mark \(\beta=1\).}
\label{fig:beta_prediction_errors}
\end{figure}

Figure~\ref{fig:beta_prediction_errors} shows that \(\beta=0\) increases five-step error and reduces closed-loop success relative to \(\beta=1\) in all nine settings.
This joint degradation is consistent with prediction accuracy playing a role under strong attenuation.
Under partial attenuation, the relationship varies across checkpoints.
For example, reducing \(\beta\) from \(1\) to \(0.5\) in the base Reacher regime increases five-step error while improving closed-loop success.
These observations motivate comparing planning performance at similar mean five-step error.

\noindent\textbf{Closed-loop planning at comparable error.}
To assess how much mean five-step error can explain the planning effects of response scaling, we compare closed-loop success after bringing the errors to similar levels.
For each task, we use one frozen base checkpoint and calibrate \(\beta=1\) and each of \(\beta=0.75\) and \(\beta=1.25\) to each other's uncalibrated five-step MSE.
Drawing on bias compensation in MPC~\citep{pannocchia2003disturbance}, we fit a constant additive offset to the terminal latent predictions on 20 of the 50 evaluation starts from seed 42.
We fix the offset before evaluation and check the \(10\%\) matching tolerance on the remaining 30 starts.
For settings that meet this tolerance, we report closed-loop success over all 50 starts.
\begin{table}[ht]
\centering
\small
\caption{Five-step MSE on 30 held-out start--goal pairs and closed-loop success (\%) on all 50 pairs from evaluation seed 42.
Each group starts with the uncalibrated result.
Indented rows calibrate that group's \(\beta\) to the uncalibrated MSE at the indicated reference scale.
A dash indicates that the held-out MSE fell outside the \(10\%\) matching tolerance.}
\label{tab:closed_loop_error_matching}
\begin{tabular}{@{}l*{4}{cc}@{}}
\toprule
& \multicolumn{2}{c}{TwoRoom}
& \multicolumn{2}{c}{PushT}
& \multicolumn{2}{c}{Reacher}
& \multicolumn{2}{c}{Cube} \\
\cmidrule(lr){2-3}
\cmidrule(lr){4-5}
\cmidrule(lr){6-7}
\cmidrule(lr){8-9}
Setting & MSE & Success & MSE & Success & MSE & Success & MSE & Success \\
\midrule
\(\beta=1\)
& 141.6 & 90 & 10.8 & 96 & 49.0 & 82 & 72.3 & 72 \\
\quad \textit{Calibrated} \(\to 0.75\)
& 141.6 & -- & 12.7 & 98 & 49.1 & 82 & 76.8 & 68 \\
\quad \textit{Calibrated} \(\to 1.25\)
& 219.4 & 90 & 15.5 & 94 & 73.5 & 88 & 81.3 & 72 \\
\addlinespace
\(\beta=0.75\)
& 97.8 & 96 & 13.5 & 94 & 47.7 & 78 & 77.3 & 80 \\
\quad \textit{Calibrated} \(\to 1\)
& 140.5 & 96 & 13.6 & -- & 49.4 & 86 & 77.3 & 74 \\
\addlinespace
\(\beta=1.25\)
& 217.9 & 86 & 15.5 & 94 & 76.8 & 74 & 83.5 & 74 \\
\quad \textit{Calibrated} \(\to 1\)
& 217.9 & -- & 15.6 & -- & 77.0 & -- & 82.9 & -- \\
\bottomrule
\end{tabular}
\end{table}

For \(\beta=0.75\), Table~\ref{tab:closed_loop_error_matching} shows that at comparable five-step MSE, the calibrated \(\beta=1\) model achieves lower closed-loop success than the uncalibrated \(\beta=0.75\) model in Cube and higher success in PushT and Reacher.
In the reverse direction, the calibrated \(\beta=0.75\) model achieves higher success than the uncalibrated \(\beta=1\) model in TwoRoom, Reacher, and Cube.
The \(1\to0.75\) calibration falls outside the matching tolerance in TwoRoom, as does \(0.75\to1\) in PushT.

For the \(\beta=1.25\) comparison, at comparable five-step MSE the calibrated \(\beta=1\) model achieves higher closed-loop success than the uncalibrated \(\beta=1.25\) model in TwoRoom and Reacher, the same success in PushT, and slightly lower success in Cube.
The reverse calibration falls outside the matching tolerance in all four tasks.

These matched comparisons suggest that mean five-step prediction error alone cannot account for the changes in closed-loop success under the \(\beta\) intervention.
\newpage
\section{Rollout Geometry}
\label{app:rollout_geometry}

To interpret the rollout response energies in Equation~\ref{eq:goal_response_energy}, we examine how small action perturbations spread the terminal predictions and change the predicted goal cost.

Fix the initial context, reference actions \(\mathbf a\), and goal \(z_g\). Let the terminal map \(\Phi_H(\mathbf a)=\hat z_{t+H}\) be differentiable with Jacobian \(\mathcal J_H\), and let \(\delta\mathbf a\) have zero mean and finite covariance \(\Sigma_{\mathbf a}\). All moments below are taken at this fixed reference.

\noindent\textbf{Terminal response covariance.}
The first-order terminal response and covariance-weighted Jacobian are
\[
\delta z^{(1)}=D\Phi_H(\mathbf a)[\delta\mathbf a]=\mathcal J_H\delta\mathbf a,
\qquad
\widetilde{\mathcal J}_H=\mathcal J_H\Sigma_{\mathbf a}^{1/2},
\]
where \(\Sigma_{\mathbf a}^{1/2}\) is the symmetric positive semidefinite square root. Let \(e=\Phi_H(\mathbf a)-z_g\ne0\) and let \(u\) be the unit vector from the terminal prediction toward the goal. Define
\[
S_{\mathrm{total}}=\lVert\widetilde{\mathcal J}_H\rVert^2,
\qquad
S_{\parallel}=\lVert\widetilde{\mathcal J}_H^\top u\rVert^2,
\qquad
S_{\perp}=S_{\mathrm{total}}-S_{\parallel}.
\]
Since \(\mathbb E[\delta z^{(1)}]=0\), its covariance is
\begin{equation}
\operatorname{Cov}(\delta z^{(1)})
=\mathcal J_H\mathbb E[\delta\mathbf a\delta\mathbf a^\top]\mathcal J_H^\top
=\widetilde{\mathcal J}_H\widetilde{\mathcal J}_H^\top.
\label{eq:app_rollout_covariance}
\end{equation}
Taking the trace and projecting along \(u\) and its orthogonal complement gives
\begin{equation}
\begin{aligned}
\mathbb E\lVert\delta z^{(1)}\rVert^2
&=\operatorname{tr}\!\left(\widetilde{\mathcal J}_H\widetilde{\mathcal J}_H^\top\right)=S_{\mathrm{total}},\\
\mathbb E\bigl[(u^\top\delta z^{(1)})^2\bigr]
&=u^\top\widetilde{\mathcal J}_H\widetilde{\mathcal J}_H^\top u=S_{\parallel},\\
\mathbb E\lVert(I-uu^\top)\delta z^{(1)}\rVert^2
&=S_{\mathrm{total}}-S_{\parallel}=S_{\perp}.
\end{aligned}
\label{eq:app_rollout_response_energies}
\end{equation}
The last equality uses the orthogonality of the two projections.

\noindent\textbf{First-order variation in goal cost.}
For \(c(\mathbf a)=\lVert\Phi_H(\mathbf a)-z_g\rVert^2\), the chain rule gives \(\nabla_{\mathbf a}c=2\mathcal J_H^\top e\). Hence its first-order variation is
\begin{equation}
\delta c^{(1)}=Dc(\mathbf a)[\delta\mathbf a]
=2e^\top\mathcal J_H\delta\mathbf a
=-2\lVert e\rVert u^\top\delta z^{(1)}.
\label{eq:app_goal_cost_variation}
\end{equation}
Using Equation~\ref{eq:app_rollout_response_energies},
\begin{equation}
\operatorname{Var}(\delta c^{(1)})
=(\nabla_{\mathbf a}c)^\top\Sigma_{\mathbf a}\nabla_{\mathbf a}c
=4\lVert e\rVert^2S_{\parallel}.
\label{eq:app_goal_cost_variance}
\end{equation}
Thus, at a fixed goal distance, \(S_{\parallel}\) determines the variance of the first-order change in predicted goal cost.
The ratio \(q\) measures how the balance between goal-direction and orthogonal response energies changes relative to the base model.

\noindent\textbf{Linear dynamics as a special case.}
For linear controlled dynamics,
\begin{equation}
\hat z_{j+1}=M\hat z_j+Ba_j,
\qquad F(z,a)=(M-I)z+Ba.
\label{eq:app_linear_controlled_dynamics}
\end{equation}
Unrolling gives \(\Phi_H(\mathbf a)=M^H\hat z_t+\sum_{k=0}^{H-1}M^{H-1-k}Ba_{t+k}\), so differentiation yields
\begin{equation}
\mathcal J_H
=\begin{bmatrix}M^{H-1}B&M^{H-2}B&\cdots&B\end{bmatrix}.
\label{eq:app_linear_controllability_map}
\end{equation}
This recovers the finite-horizon controllability map used in the linear planning analysis~\citep{wang2026temporal}. Since \(\Phi_H(\mathbf a+\delta\mathbf a)-\Phi_H(\mathbf a)=\mathcal J_H\delta\mathbf a\) holds exactly, Equations~\ref{eq:app_rollout_covariance} and \ref{eq:app_rollout_response_energies} also describe finite terminal changes in this special case. The squared goal cost still contains a quadratic term in finite action perturbations, so Equation~\ref{eq:app_goal_cost_variance} concerns its first-order variation.

\section{Generalization Across Planners and World Models}
\label{app:rollout_generalization}

\begin{figure}[ht]
\centering
\includegraphics[width=\linewidth]{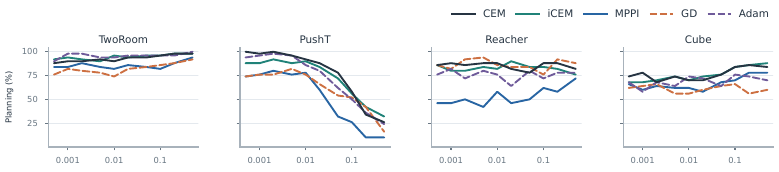}
\caption{LeWM planning success across curvature regularization strengths.
Solid lines show the sampling-based solvers and dashed lines show the gradient-based solvers.}
\label{fig:solver_sweep}
\end{figure}

\noindent\textbf{Planning across solvers on LeWM.}
To test whether the planning trends in Figure~\ref{fig:propagation_lambda} carry over to other solvers, we further evaluate LeWM with four other solvers for each task and regularization strength. 
Within each task, they are evaluated on the same 50 episodes sampled with evaluation seed 42.
The improved cross-entropy method (iCEM) uses temporally correlated sampling and reuses elite candidates~\citep{pinneri2021sample}. Model predictive path integral control (MPPI) updates the action sequence using cost-weighted perturbations~\citep{williams2017model}. Gradient descent (GD) updates the action sequence along the negative gradient of the planning cost~\citep{bonnans2006numerical}. Adam adapts these updates using estimates of the first and second moments of the gradients~\citep{kingma2014adam}. 
Appendix~\ref{app:experimental_settings} gives the solver settings.
Figure~\ref{fig:solver_sweep} shows that the main trends are broadly consistent across solvers.

\noindent\textbf{Rollout geometry on DINO-WM.}
We next test whether the rollout-geometry trends extend to DINO-WM~\citep{zhou2024dino}.
Following the temporal straightening setup~\citep{wang2026temporal}, we apply the diagnostics from Section~\ref{sec:6} to all four tasks using five-step rollouts and a zero-action reference sequence.
DINO-WM's terminal cost sums the mean squared distances to the visual and proprioceptive goal embeddings.
To match this cost, we divide each block of the rollout Jacobian and terminal goal displacement by the square root of the corresponding embedding dimension before computing \(R_{\mathrm{total}}\) and \(q\).

Table~\ref{tab:dinowm_generalization} shows that at \(\lambda=0.1\), DINO-WM exhibits overall contraction and greater relative goal-direction retention across all four tasks, consistent with the trend observed in LeWM.

\begin{table}[ht]
\centering
\small
\caption{Rollout geometry and CEM planning success for DINO-WM across four tasks. The geometry diagnostics use each task's \(\lambda=0\) checkpoint as the reference. CEM success rates (\%) are measured over 50 closed-loop episodes per setting.}
\label{tab:dinowm_generalization}
\begin{tabular}{@{}r*{4}{ccc}@{}}
\toprule
& \multicolumn{3}{c}{TwoRoom} & \multicolumn{3}{c}{PushT}
& \multicolumn{3}{c}{Reacher} & \multicolumn{3}{c}{Cube} \\
\cmidrule(lr){2-4}\cmidrule(lr){5-7}\cmidrule(lr){8-10}\cmidrule(lr){11-13}
\(\lambda\)
& \(R_{\mathrm{total}}\) & \(q\) & CEM
& \(R_{\mathrm{total}}\) & \(q\) & CEM
& \(R_{\mathrm{total}}\) & \(q\) & CEM
& \(R_{\mathrm{total}}\) & \(q\) & CEM \\
\midrule
\(0\)    & \(1.00\) & \(1.00\) & 100 & \(1.00\) & \(1.00\) & 42 & \(1.00\) & \(1.00\) & 54 & \(1.00\) & \(1.00\) & 86 \\
\(0.01\) & \(0.27\) & \(2.37\) & 100 & \(1.03\) & \(1.58\) & 44 & \(0.57\) & \(5.16\) & 62 & \(0.12\) & \(6.48\) & 70 \\
\(0.1\)  & \(0.17\) & \(3.83\) & 100 & \(0.13\) & \(3.75\) & 66 & \(0.31\) & \(7.63\) & 66 & \(0.13\) & \(6.95\) & 74 \\
\bottomrule
\end{tabular}
\end{table}

\newpage
\section{Additional Encoder Attention Examples}
\label{app:attention_map}

The following figures extend the qualitative comparison in Section~\ref{sec:attention} using additional observation sequences. 
At each task's beneficial regimes, we choose the checkpoint with the highest CEM success among those that outperform their base counterpart from the same training seed.
Within each sequence, all checkpoints are compared on the same recorded frames at offsets \(t=0,2,4\) from the start of the displayed window.
Colors are normalized independently within each panel to indicate relative spatial emphasis. 
We interpret these patterns in relation to each task's
goal and environmental constraints.

\begin{figure}[ht]
\centering
\includegraphics[width=\linewidth]{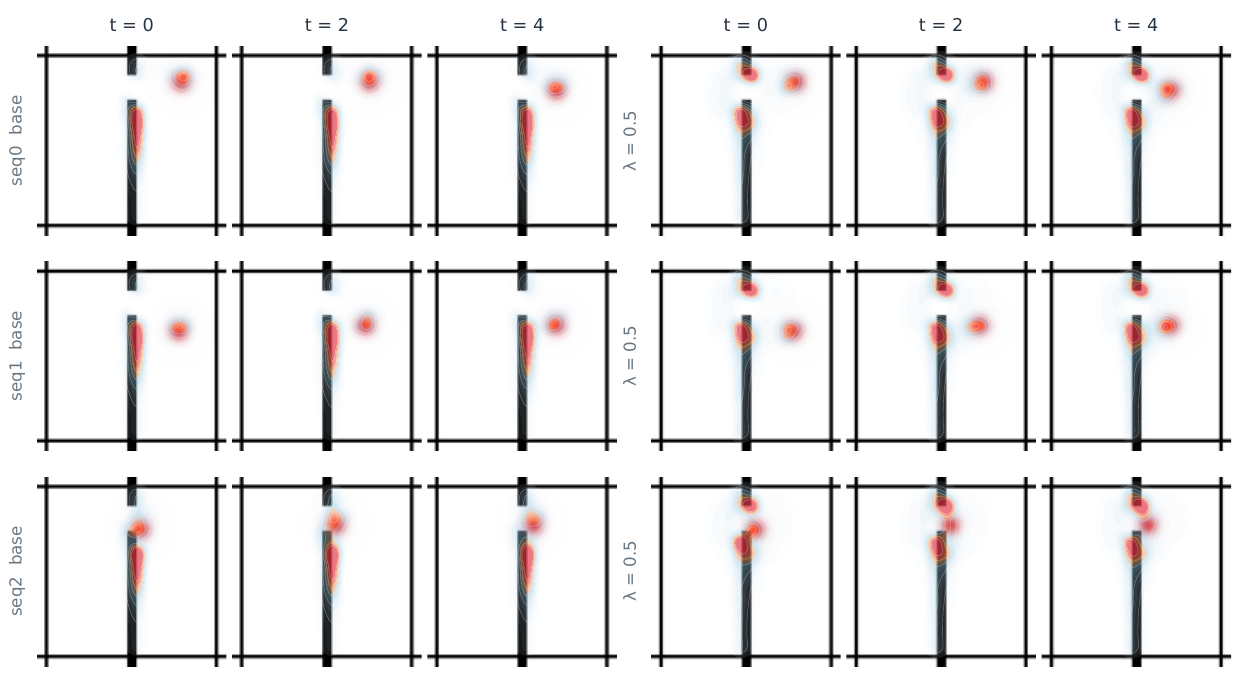}
\end{figure}
\noindent\textbf{TwoRoom}
shows attention around the partition and the wall edges bordering the passage, in addition to the agent's neighborhood. The passage boundaries remain prominent in the regularized maps. For goals in the other room, these boundaries determine which routes the agent can take. Attention to this structure is therefore consistent with representing the spatial constraints on goal-reaching.

\newpage
\begin{figure}[ht]
\centering
\includegraphics[width=\linewidth]{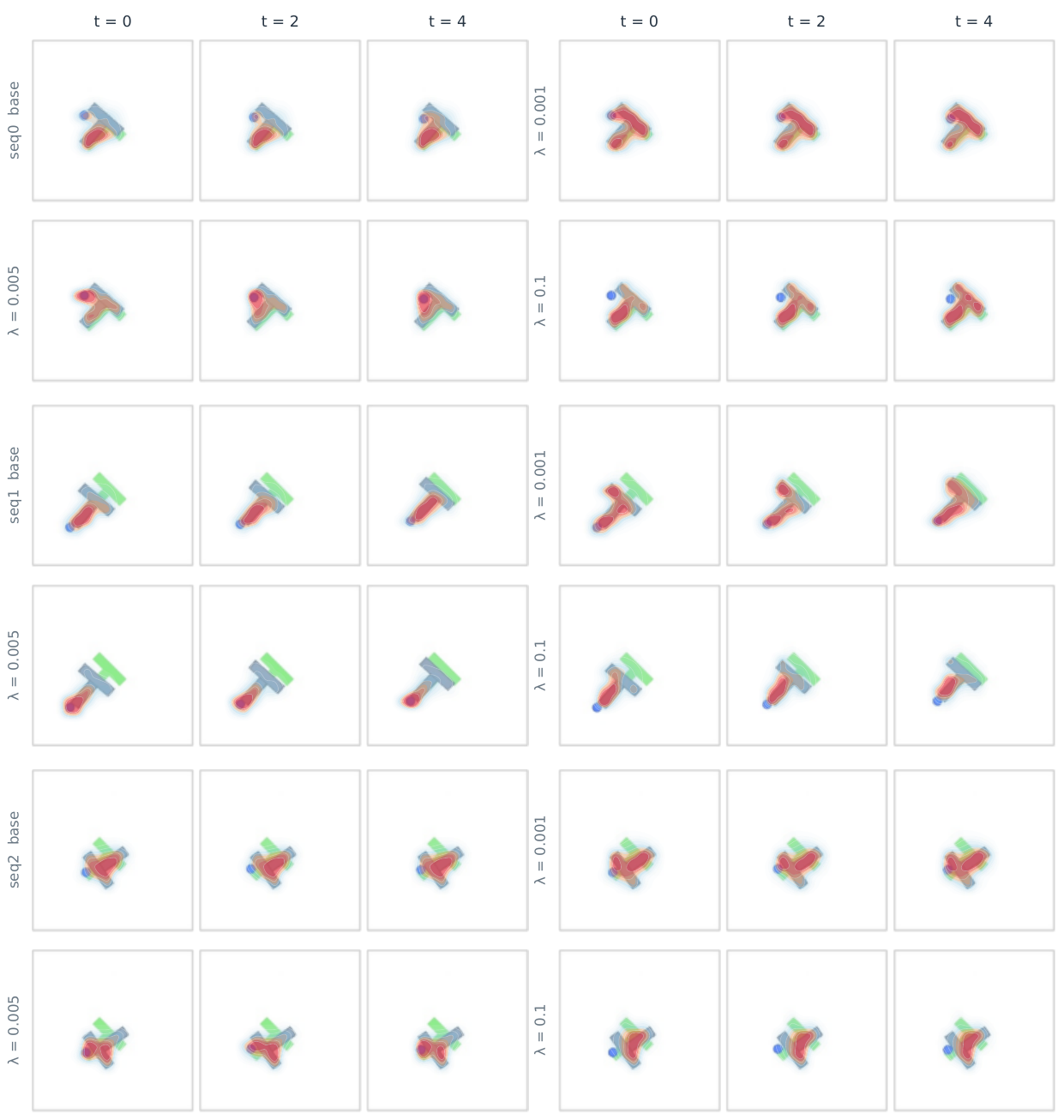}
\end{figure}
\noindent\textbf{PushT}
shows more pronounced attention around the circular pusher at \(\lambda=0.001\) than in the base maps, alongside attention to the T-shaped object. Their relative positions affect how contact moves and rotates the object, making both relevant to reaching the desired configuration. To illustrate how this pattern changes under stronger regularization, we also show an intermediate setting at \(\lambda=0.005\) and the over-regularized setting at \(\lambda=0.1\). The pusher remains prominent at \(\lambda=0.005\). At \(\lambda=0.1\), hotspots remain on parts of the T-shaped object, with less emphasis on the pusher in the displayed frames. This shift accompanies the decline in planning success at this strength.

\newpage
\begin{figure}[ht]
\centering
\includegraphics[width=\linewidth]{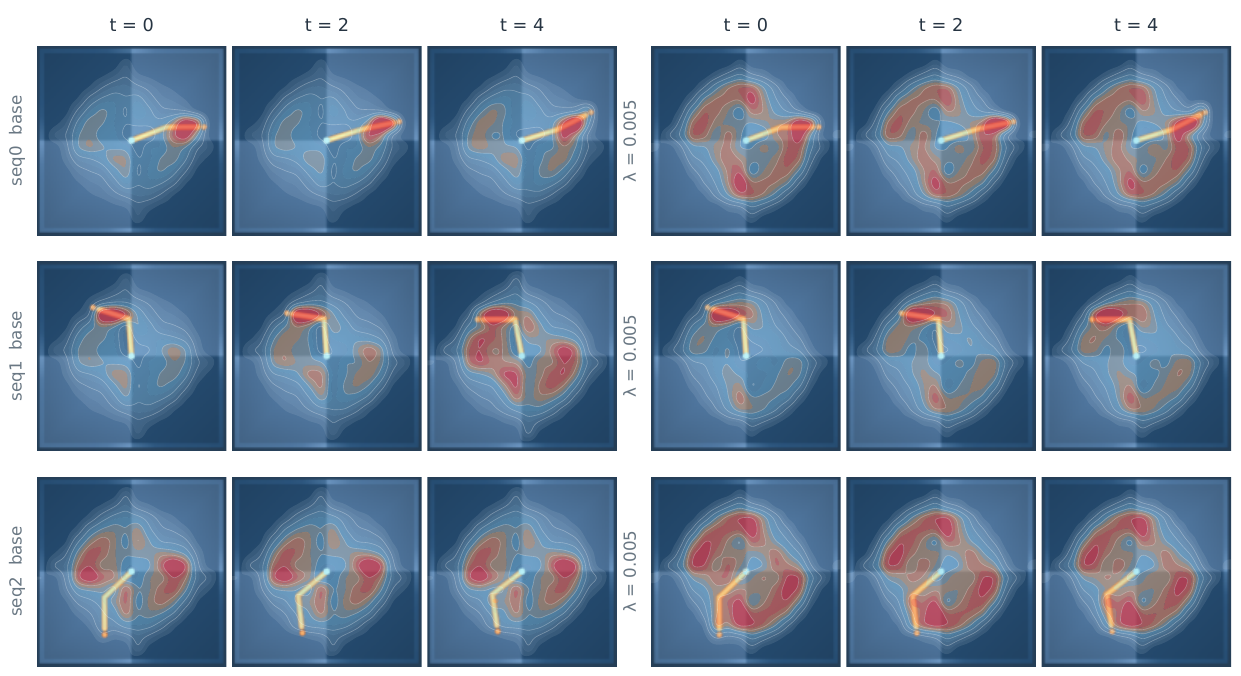}
\end{figure}
\noindent\textbf{Reacher}
shows attention on parts of the arm in both checkpoints. When the arm is extended, the regularized maps retain a hotspot on the distal link and give more emphasis to the surrounding workspace. In this task, the goal is to match a target joint configuration, making the arrangement of the links relevant to goal-reaching.

\begin{figure}[ht]
\centering
\includegraphics[width=\linewidth]{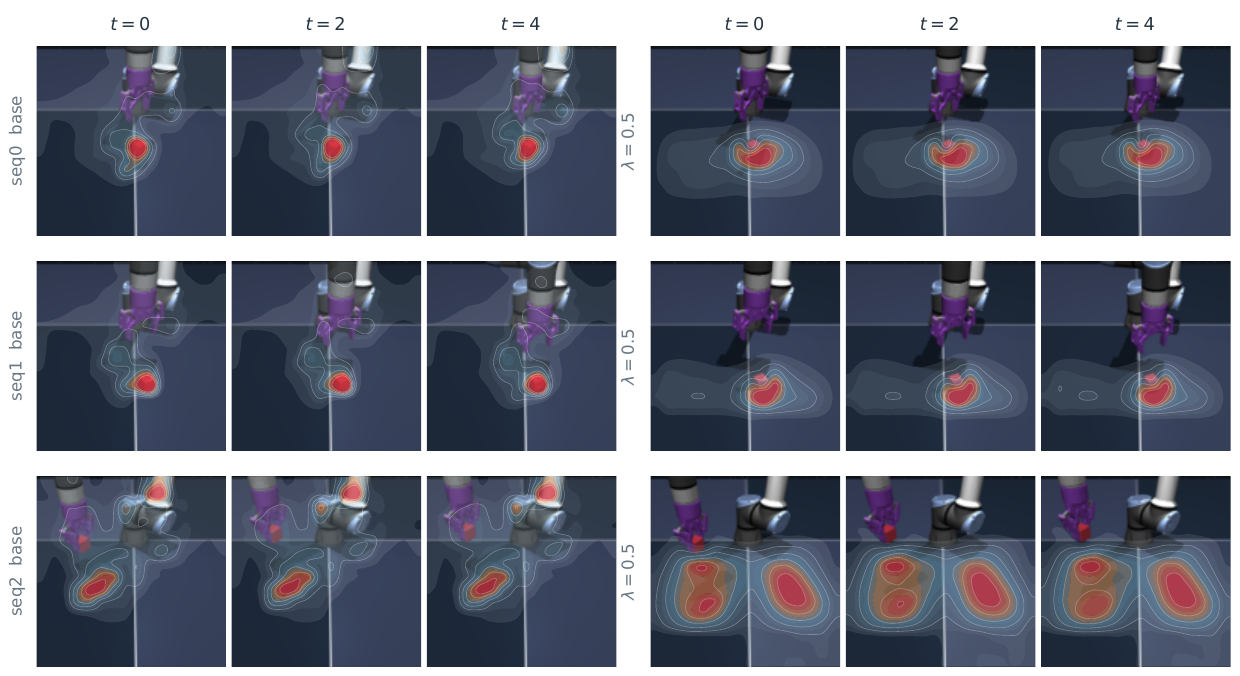}
\end{figure}
\noindent\textbf{Cube}
shows attention extending over the tabletop near and below the cube, while hotspots on the upper robot arm become less prominent. The task requires moving the cube to a target position, so the spatial relationship between the cube and its immediate workspace is relevant to planning.

\end{document}